*Article*

# Automatic Ship Detection of Remote Sensing Images from Google Earth in Complex Scenes Based on Multi-Scale Rotation Dense Feature Pyramid Networks

**Xue Yang [1, 2], Hao Sun[1], Kun Fu[1, 2, *], Jirui Yang[1, 2], Xian Sun[1], Menglong Yan[1], and Zhi Guo[1]**

[1]   Key Laboratory of Technology in Geo-spatial Information Processing and Application System, Institute of Electronics, Chinese Academy of Sciences, Beijing, China; sun.010@163.com (H. S.); sunxian@mail.ie.ac.cn (X. S.); yanmenglong@foxmail.com (M. Y.); guozhi@mail.ie.ac.cn (Z. G.)

[2]   School of Electronic, Electrical and Communication Engineering, University of Chinese Academy of Sciences, Beijing, China; yangxue0827@126.com (X. Y.); yang_jirui@sina.com (J. Y.)

*   Correspondence: kunfuiecas@gmail.com (K. F.); Tel.: +010-58887208-8273

**Abstract:** Ship detection has been playing a significant role in the field of remote sensing for a long time but it is still full of challenges. The main limitations of traditional ship detection methods usually lie in the complexity of application scenarios, the difficulty of intensive object detection and the redundancy of detection region. In order to solve such problems above, we propose a framework called Rotation Dense Feature Pyramid Networks (R-DFPN) which can effectively detect ship in different scenes including ocean and port. Specifically, we put forward the Dense Feature Pyramid Network (DFPN), which is aimed at solving the problem resulted from the narrow width of the ship. Compared with previous multi-scale detectors such as Feature Pyramid Network (FPN), DFPN builds the high-level semantic feature-maps for all scales by means of dense connections, through which enhances the feature propagation and encourages the feature reuse. Additionally, in the case of ship rotation and dense arrangement, we design a rotation anchor strategy to predict the minimum circumscribed rectangle of the object so as to reduce the redundant detection region and improve the recall. Furthermore, we also propose multi-scale ROI Align for the purpose of maintaining the completeness of semantic and spatial information. Experiments based on remote sensing images from Google Earth for ship detection show that our detection method based on R-DFPN representation has a state-of-the-art performance.

---

## 1. Introduction

With the development of remote sensing technology, more and more attention has been paid to the research of remote sensing images. Ship detection has been playing an important role in the field of remote sensing for a long time and can promote the national defense construction, port management, cargo transportation, and maritime rescue. Although many ship detection methods have been proposed before, this task still poses a great challenge due to the existence of uncertainties such as light, disruptors, density of the ship and so on.

In the past few years, some traditional methods have been proposed for ship detection [1-5]. Some methods adopt the following ideas: Firstly, the sea-land segmentation is carried out through the features of the texture and the shape, and extracts the sea region as region of interest (ROI). Then an algorithm such as contrast box algorithm [6], semi-supervised hierarchical classification [7] is used to get the candidate object region. Finally, filtering false box by post processing to get the final detection results. Bi F et al. [8] uses a bottom-up visual attention mechanism to select prominent candidate regions throughout the detection scene. Although these methods have shown promising performance, they have poor practicability in complex scenarios.

With the application of deep convolutional networks [9-15] in object detection, more and more efficient detection algorithms are proposed, such as region proposals with CNNs (RCNN) [16], Spatial Pyramid Pooling Network (SSP-Net) [17] and Fast-RCNN [18]. Faster-RCNN [19] proposes Region Proposal Network (RPN) structure and improves the detection efficiency while achieving end-to-end training. Instead of relying on regional proposals, You Only Look Once (YOLO) [20] and Single Shot MultiBox Detector (SSD) [21] directly estimate objects region and truly enable real-time detection. Feature Pyramid Network (FPN) [22] adopts the multi-scale feature pyramid form and makes full use of the feature map to achieve better detection results. Region-based Fully Convolutional Networks (R-FCN) [23] builds a fully convolution network, which greatly reduces the number of parameters, improves the detection speed, and has a good detection effect.

These visual detection algorithms above are also widely used in remote sensing ship detection. Zhang R et al. [24] proposes a new method of ship detection based on convolution neural network (SCNN), combined with an improved saliency detection method. Kang M et al. [25] take the objects proposals generated by Faster R-CNN for the guard windows of CFAR algorithm, then pick up the small-sized targets, thus reevaluating the bounding boxes which have relative low classification scores in detection network. Liu Y et al. [26] presents a framework of Sea-Land Segmentation-based Convolutional Neural Network (SLS-CNN) for ship detection that attempts to combine the SLS-CNN detector, saliency computation and corner features. These methods above are known as horizontal region detection. However, in real life, for a ship with a large aspect ratio, once the angle is inclined, the redundancy region will be relatively large, and it is unfavorable to the operation of non-maximum suppression, often resulting in missing detection, as shown in Figure 1. In order to solve the same problem, Jiang Y et al. [27] proposed the Rotational Region CNN (R²CNN) and achieved outstanding results on scene text detection. However, since R²CNN still uses horizontal anchors at the first stage, the negative effects of non-maximum suppression still exist. RRPN [28] uses rotation anchors which effectively improve the quality of the proposal. However, it has a serious problem of information loss when processing the ROI, resulting in a much lower detection indicator than the R²CNN.

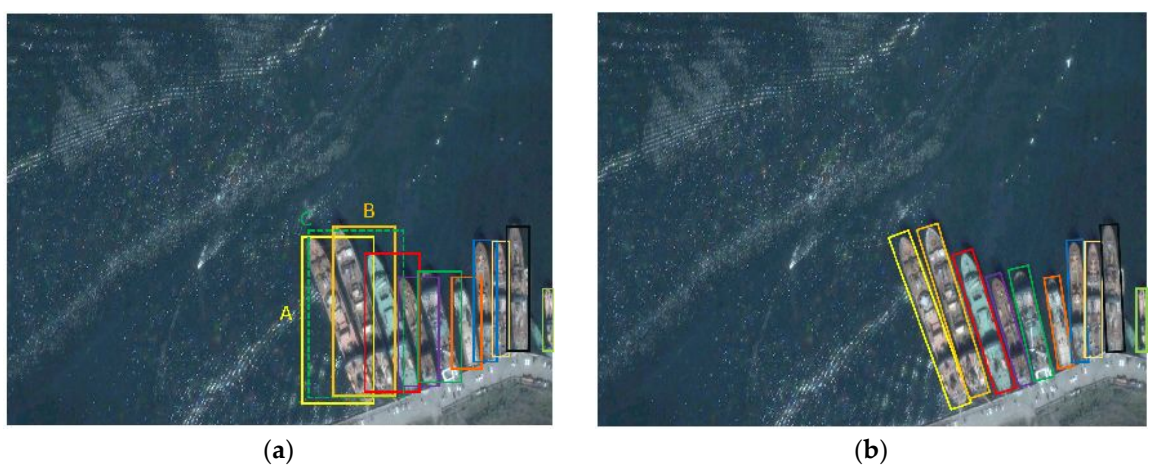

(**a**)                                                                    (**b**)

**Figure 1.** (**a**) Horizontal region detection with large redundancy region, bounding box A and B merge into C in the final prediction. (**b**) Rotation region detection with fitting detection region.

This paper presents an end-to-end detection framework called Rotation Dense Feature Pyramid Networks (R-DFPN) to solve problems above. The framework is based on a multi-scale detection network [29-31], using a dense feature pyramid network, rotation anchors, multi-scale ROI Align and other structures. Compared with other rotation region detection methods such as RRPN [28] and R²CNN [27], our framework is more suitable for ship detection tasks, and has achieved the state-of-the-art performance. The main contributions of this paper include:

1. Different from previous detection models, we build a new ship detection framework based on rotation region which can handle different complex scenes, detect intensive objects and reduce redundant detection region.

2. We propose the feature pyramid of dense connections based on a multi-scale detection framework, which enhances feature propagation, encourages feature reuse and ensures the effectiveness of detecting multi-scale objects.

3. We adopt rotation anchors to avoid side effects of non-maximum suppression and overcome the difficulty of detecting densely arranged targets, and eventually get a higher recall.

4. We use multi-scale ROI Align to solve the problem of feature misalignment instead of ROI pooling, and to get the fixed-length feature and regression bounding box to fully keep the completeness of semantic and spatial information through the horizontal circumscribed rectangle of proposal.

Experiments based on remote sensing images from Google Earth for ship detection show that our detection method based on R-DFPN representation has a state-of-the-art performance. The rest of this paper is organized as follows. Section 2 introduces the details of the proposed method. Section 3 presents experiments conducted on remote sensing dataset to validate the effectiveness of the proposed framework. Section 4 discusses the results of the proposed method. Finally, Section 5 concludes this paper.

## 2. Proposed Method

In this section we will detail the various parts of the R-DFPN framework. Figure 2 shows the overall framework of R-DFPN. The framework mainly consists of two parts：Dense Feature Pyramid Network (DFPN) for feature fusion and Rotation Region Detection Network (RDN) for prediction. Specifically, DFPN can generate feature maps that are fused by multi-scale features for each input image. Then getting rotational proposals from the RPN to provide high-quality region proposals for the next stage. Finally, the location regression and class prediction of proposals are processed in the Fast-RCNN stage.

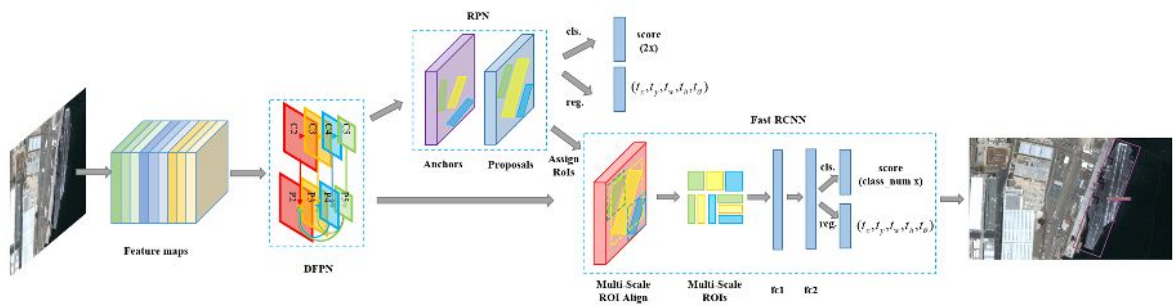

**Figure 2.** Overall framework of R-DFPN.

### 2.1. DFPN

As we all know, low-level feature semantic information is relatively few, but the object location is accurate. On the contrary, high-level feature semantic information is rich, but the object location is relatively rough. The feature pyramid is an effective multi-scale method to fuse multi-level information. Feature Pyramid Networks (FPN) achieved very good results in small object detection tasks. It uses the feature pyramid, which is connected via top-down pathway and lateral connection.

Ship detection can be considered a task to detect small objects because of the characteristic of large aspect ratio of ship. Meanwhile, considering the complexity of background in remote sensing images, there are a lot of ship-like interferences in the port such as roofs, container piles and so on. Therefore, the feature information obtained through the FPN may be not sufficient to distinguish these objects. In order to solve the problems above, we design Dense Feature Pyramid Network (DFPN), which uses a dense connection, enhances feature propagation and encourages feature reuse [32].

Figure 3 shows the architecture of DFPN based on ResNets [33]. In the bottom-up feedforward network, we still choose multi-level feature maps as $\{C_2, C_3, C_4, C_5\}$, corresponding to the last layer of each residual block which have strong semantic features. We note that they have strides of $\{4, 8, 16, 32\}$ pixels. In the top-down network, we get higher resolution features by lateral connections and dense connections as $\{P_2, P_3, P_4, P_5\}$. For example, in order to get $P_2$, we first reduce the number of $C_2$ channels by using a 1x1 convolutional layer, then we use nearest neighbor up-sampling for all the preceding feature maps. We merge them by concatenating rather than simply adding. Finally, we eliminate the aliasing effects of up-sampling through a 3x3 convolutional layer, while reducing the number of channels. After the iteration above, we get the final feature maps $\{P_2, P_3, P_4, P_5\}$.

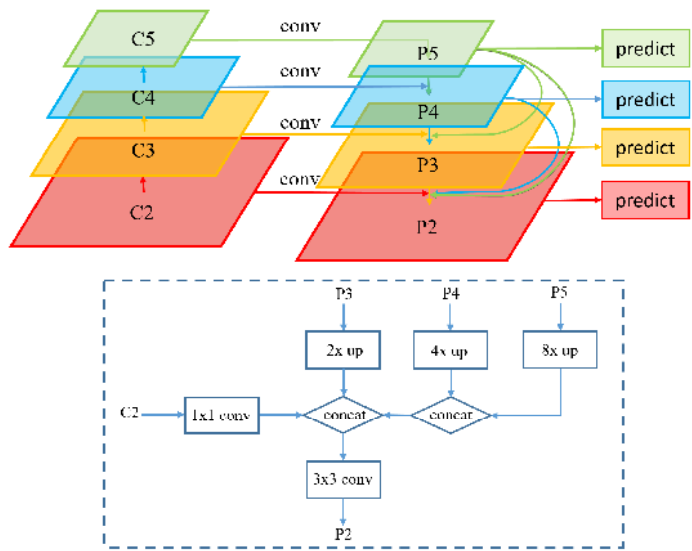

**Figure 3.** A multi-scale feature pyramid connection. Each feature map is densely connected, merged by concatenating.

It should be noted that we do not use shared classification and regression for the feature pyramid. We believe that this can make each feature map perform better and generate more information. In order to reduce the number of parameters, we set the number of channels for all feature maps to 256 at the same time.

Through a large number of experimental comparison, we find that the use of DFPN can significantly improve the detection performance due to the smooth feature propagation and feature reuse.

## 2.2 RDN

Similar to the traditional detection framework, Rotation Region Detection Network also contains two stages: RPN and Fast-RCNN. In order to achieve the detection of rotation objects, two stages above have to make changes, as shown in Figure 2. In the RPN stage, we have to redefine the representation of the rectangle to get the "Rotation Bounding Box" at first. After that we generate rotation proposals by regressing the rotation anchors to reduce the impact of non-maximum suppression and improve the recall. Then each proposal obtains a fixed-length feature vector through the Multi-scale ROI Align layer to preserve the complete feature information. In order to match the operation of ROI Align, we regress the horizontal circumscribed rectangle of proposal instead of itself in the second stage. In addition, through two fully connected layers, we conduct a position prediction and classification. Finally, the final result is obtained by non-maximum suppression.

### 2.2.1. Rotation Bounding Box

The traditional bounding box is a horizontal rectangle, so its representation is relatively simple, using $(x_{\min}, y_{\min}, x_{\max}, y_{\max})$ representation. It respectively represents the coordinates of the upper left and lower right corners of the bounding box. But this is obviously no longer suitable for representing a rotation bounding box. In order to represent the bounding box more generally, we use the five variables $(x, y, w, h, \theta)$ to uniquely determine the arbitrary bounding box. As shown in Figure 4, x and y represent the coordinates of the center point. Rotation angle $(\theta)$ is the angle at which the horizontal axis (x-axis) rotates counterclockwise to the first edge of the encountered rectangle. At the same time we define this side is width, the other is height. We note that the range of angles is $[-90, 0)$.

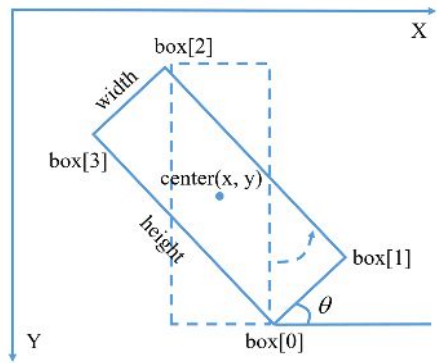

**Figure 4.** General representation of bounding box.

### 2.2.2. Rotation Anchor/Proposal

In contrast to R²CNN, which still uses horizontal anchors in detecting scene text, we use a rotation anchors at the RPN stage. For a ship with large aspect ratio, it is likely that a horizontal proposal contains more than one ship after non-maximum suppression, resulting in missing detection. In this paper, we use the three parameters of scale, ratio and angle to generate anchors. Taking into account of the characteristics of the ship, the ratio of $\{1:3, 3:1, 1:5, 5:1, 1:7, 7:1, 1:9, 9:1\}$ was adopted. Then we assign a single scale to each feature map, the size of the scale is $\{50, 150, 250, 350, 500\}$ pixels on $\{P_2, P_3, P_4, P_5, P_6\}$ respectively. Then we add six angles $\{-15°, -30°, -45°, -60°, -75°, -90°\}$ to control the orientation so as to cover the object more effectively. Each feature point for each feature map will generate 48 anchors $(1 \times 8 \times 6)$, 240 outputs $(5 \times 48)$ for each regression layer and 96 outputs $(2 \times 48)$ for each classification layer. Figure 5 shows excellent results using a multi-scale framework.

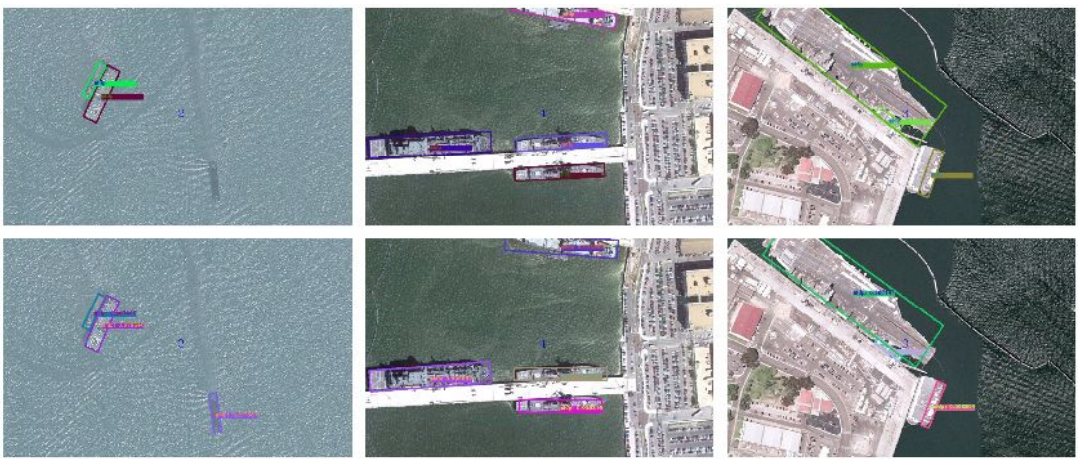

**Figure 5.** Multi-scale detection results. First row: ground-truth (some small objects are not labeled, such as first column). Second row: detection results of R-DFPN.

## 2.2.3. Non-Maximum Suppression

Intersection-over-Union (IoU) computation is a core part of non-maximal suppression. However, the rotation proposals can be generated in any orientations, so IoU computation on axis aligned proposals may lead to an inaccurate IoU of skew interactive proposals and further ruin the proposal learning. An implementation for Skew IoU computation [28] with thought of triangulation is proposed to deal with this problem. We need to use non-maximum suppression twice during the entire training process, the first is to get the appropriate proposals, and the second is the post-processing of the predictions. In traditional horizontal region detection tasks, the non-maximum suppression of both stages encounters such difficulty that once the objects are densely arranged, some proposals or predictions will be discarded because of large overlap, resulting in missing detection, as shown in Figure 6 (a) and Figure 6 (b). Too many redundant regions in the horizontal rectangle lead to these undesirable results, while the rotation region detection avoids this problem, as shown in Figure 6 (c) and Figure 6 (d).

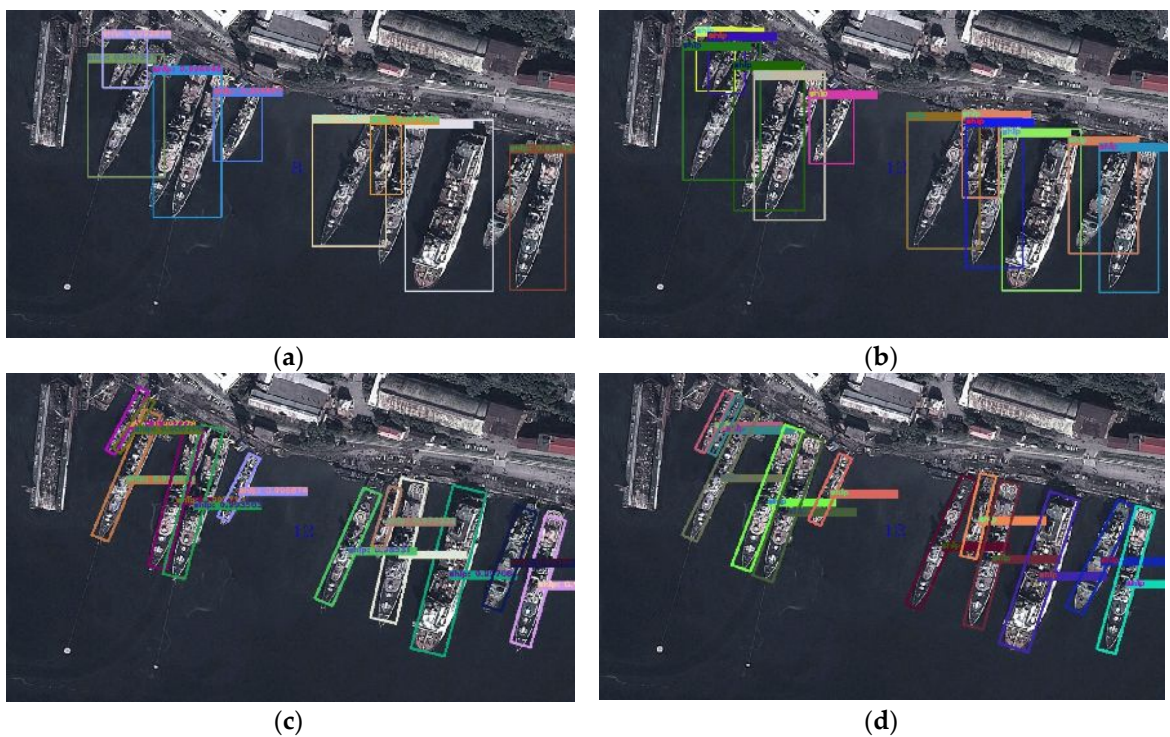

**Figure 6.** (**a**) Detection result of horizontal region detection (appear missing detection, because of the non-maximum suppression); (**b**) Horizontal ground-truth; (**c**) Detection result of rotation region detection; (**d**) Rotation ground-truth.

## 2.2.4. Multi-Scale ROI Align

RRPN uses Rotation Region-of-Interest (RROI) pooling to obtain fixed-length feature vector from proposal, which is not suitable for ship detection. Taking into account the narrow side of large aspect ratio objects and the problem of feature misalignment in ROI pooling, the final cropped ROI may not contain any useful information. Therefore, we adopted the ROI Align to process the horizontal circumscribed rectangle of the proposal to solve the problem of feature misalignment and added two pool sizes of 3:16 and 16: 3 to minimize the influence of the distortion caused by the interpolation method (no matter what the angle of the boat is, at least one of the pooling results is not seriously deformed, shown in Figure 2). In order to match the operation of the ROI Align, it is crucial that we regress the horizontal circumscribed rectangle of proposal at second stage. Figure 7 visualizes the feature cropping effect of the Multi-Scale ROI Align method.

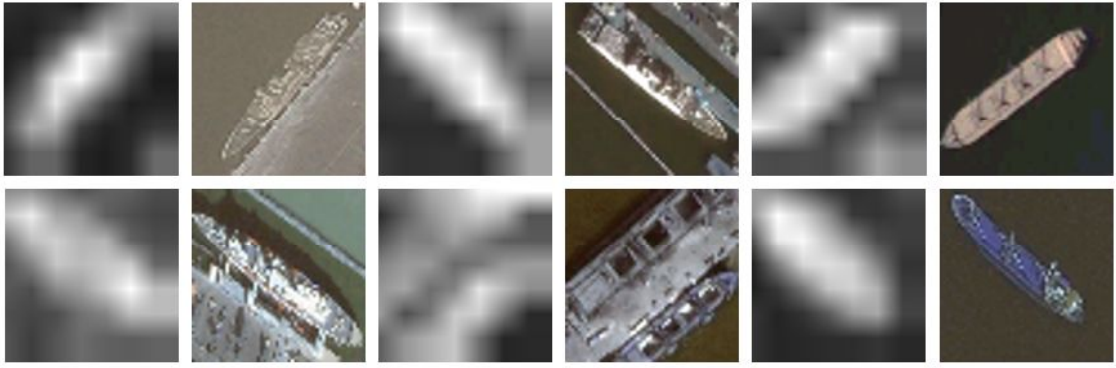

**Figure 7.** Visualization of the Multi-Scale ROI Align. Semantic and spatial information is completely preserved. Odd columns are the feature maps of the objects, and even columns are the objects.

### 2.2.5. Loss Function

During training of the RPN, each anchor is assigned a binary class label and five parametric coordinates. To train the RPN [19], we need to find positive and negative samples from all anchors, which we call a mini-batch. The positive samples anchors need to satisfy the following conditions: (i) the IoU [16] overlap between an anchor and the ground-truth is greater than 0.5, and the angular difference is less than 15 degrees, or (ii) an anchor has the highest IoU overlap with a ground-truth. Negative samples are defined as: (i) IoU overlap less than 0.2, or (ii) IoU overlap greater than 0.5 but the angular difference is greater than 15 degrees. Anchors that are neither positive nor negative are discarded. We use multi-task loss to minimize the objective function, which is defined as follows:

$$L(p_i, l_i, t_i^*, t_i) = \frac{1}{N_{cls}}\sum_i L_{cls}(p_i, l_i) + \lambda \frac{1}{N_{reg}}\sum_i p_i L_{reg}(t_i^*, t_i) \tag{1}$$

where $l_i$ represents the label of the object, $p_i$ is the probability distribution of various classes calculated by the softmax function, $t_i$ represents the predicted five parameterized coordinate vectors, $t_i^*$ represents the offset of ground-truth and positive anchors. The hyper-parameter $\lambda$ in Eq. 1 controls the balance between the two task losses and all experiments use $\lambda=1$ in this paper. In addition, the functions $L_{cls}$ and $L_{reg}$ [18] are defined as:

$$L_{cls}(p, l) = -\log pl \tag{2}$$

$$L_{reg}(t_i^*, t_i) = smooth_{L_1}(t_i^* - t_i) \tag{3}$$

$$smooth_{L_1}(x) = \begin{cases} 0.5x^2, if\ |x|<1 \\ |x|-0.5, otherwise \end{cases} \tag{4}$$

The parameterized coordinate regression mode is as follows:

$$\begin{aligned} t_x &= (x-x_a)/w_a, t_y = (y-y_a)/h_a, \\ t_w &= \log(w/w_a), t_h = \log(h/h_a), \\ t_\theta &= \theta - \theta_a + k\pi/2 \end{aligned} \tag{5}$$

$$\begin{aligned} t_x^* &= (x^*-x_a)/w_a, t_y^* = (y^*-y_a)/h_a, \\ t_w^* &= \log(w^*/w_a), t_h = \log(h^*/h_a), \\ t_\theta^* &= \theta^* - \theta_a + k\pi/2 \end{aligned} \tag{6}$$

where $x, y, w$ and $h$ denote the box's center coordinates and its width and height. Variables $x$, $x_a$ and $x^*$ are for the predicted box, anchor box, and ground-truth box respectively (likewise for $y, w, h$). The parameter $k \in Z$ to keep $\theta$ in the range $[-90, 0)$. In order to keep the bounding box in the same position, w and h need to be swapped when $k$ is an odd number.

## 3. Experiments and Results

### 3.1. Implementation details

#### 3.1.1. Remote Sensing Dataset.

Our dataset is collected publicly from Google Earth with 1000 large scene images sized $16,393 \times 16,393$ pixels, covering 400 square kilometers. These satellite remote sensing images have the red, green and blue tri-band information after geometric correction. Their format is geotif with latitude and longitude information. The images contain scenes of civilian ports, military bases, offshore area and far seas. We divide the images into 600x1000 sub-images with an overlap 0.1, then filter out images that do not contain ships, and result with 8,000 images in final. The ratio of training set to test set is 1: 4. In the training process, we flip the image randomly, while subtracting the mean value $[103.939, 116.779, 123.68]$.

#### 3.1.2. Training

All experiments are done on the deep learning framework, Tensorlow. We use the pre-training model ResNet-101 to initialize the network. We try two training strategies: (i) alternating training. (ii) end-to-end training. Both of the strategies have the same effect. Considering it is more convenient for training with the second strategy, we adopt the second strategy in this paper. We train a total of 80k iterations, with a learning rate of 0.001 for the first 30k iterations, 0.0001 for the next 30k iterations, and 0.00001 for the remaining 20k iterations. Weight decay and momentum are respectively 0.0001 and 0.9. The optimizer chosen is MomentumOptimizer.

At the RPN stage we sample a total of 512 anchors as a mini-batch for training (256 proposals for second stage), where the ratio of positive and negative samples is 1: 1. Given that the change of angle of the ship will cause a drastic change of the IoU, we set the IoU threshold of the positive sample to 0.5 to ensure that the training has a sufficient positive sample. The feature maps are input to the RPN network through a 5x5 convolutional layer, followed by two sibling 1x1 convolution layers for regression and classification.

### 3.2. Accelerating experiment

We use millions of anchors throughout the network, but most of the anchors are worthless for a particular image and will increase the amount of computation. The operation of non-maximum suppression at first stage spends the most calculation time because it needs to select high-quality proposal from all the anchors. The experiment finds that the higher the quality of proposal, the higher its confidence score. Therefore, we can select some anchors with the highest scores at the RPN stage to perform non-maximum suppression and get the proposals we need. Figure 8 shows the impact of different combinations of anchors and proposal on the experimental results.

We find that as the number of anchors/proposals pair increases, the assessment indicators tend to be stable, but the calculation time changes drastically. We can also see from the Figure 8 that when the number of anchors/proposals pair rises to a certain amount, the result of the model shows a slight decrease. Combined with the findings above, we select the 12000 anchors with the highest confidence score, and generate 1200 proposals through the non-maximum suppression operation.

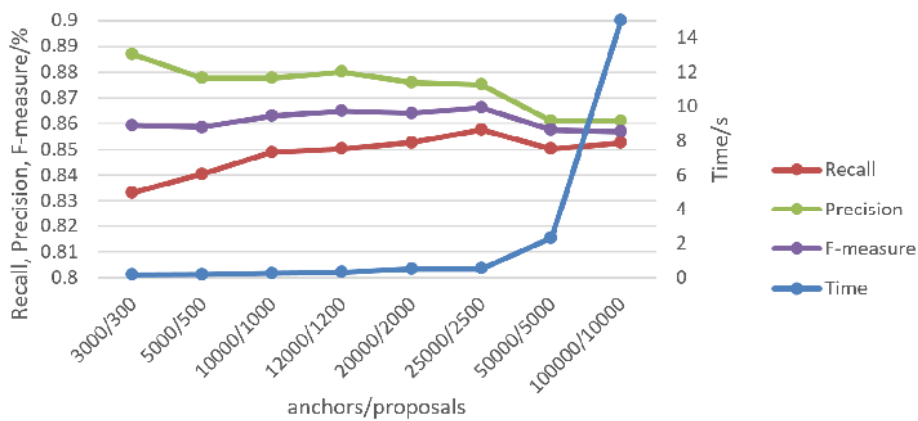

**Figure 8.** The impact of different combinations of anchors and proposal on the experimental results

### 3.3. Comparative Experiment

We perform a series of experiments on remote sensing datasets and our method achieved the state-of-the-art performance, 88.2% for Recall, 91.0% for Precision, and 89.6% for F-measure. Table 1 summarizes the experimental results of various methods, then we will compare the differences between different methods and as well as analyzing the primary role of each structure in our framework.

**Table 1.** Comparison of the performance of each detection method under the confidence score threshold of 0.5 (Based on FPN framework except Faster-RCNN). R, P, F represent Recall, Precision, and F-measure respectively.

| Detection Method | Dense Feature Pyramid | Rotation Anchor | ROI Align | Pool Size | R (%) | P (%) | F (%) |
|---|---|---|---|---|---|---|---|
| Faster | × | × | × | 7×7 | 62.7 | 96.6 | 76.0 |
| FPN | × | × | × | 7×7 | 75.5 | **97.7** | 85.2 |
| RRPN | × | √ | × | 7×7 | 68.8 | 71.1 | 69.9 |
| R²CNN | × | × | × | 7×7, 16×3, 3×16 | 80.8 | 88.7 | 84.6 |
| R-DFPN-1 | × | √ | × | 7×7 | 82.6 | 86.6 | 84.5 |
| R-DFPN-2 | √ | √ | × | 7×7 | 84.7 | 88.8 | 86.7 |
| R-DFPN-3 | √ | √ | × | 7×7, 16×3, 3×16 | 85.7 | 88.1 | 86.9 |
| R-DFPN-4 | √ | √ | √ | 7×7, 16×3, 3×16 | **88.2** | 91.0 | **89.6** |

Faster-RCNN and FPN have poor Recall due to non-maximum suppression. When the ships are arranged densely, the horizontal region detection method seems powerless. RRPN uses the RROI Pooling method, which has the problem of feature misalignment. The loss of semantic and spatial information has a huge impact on narrow ship detection. R2CNN generates horizontal proposals at the first stage, so the feature information loss is not serious, but the quality of the proposals is not high due to the non-maximum suppression, while missed detection rate is still high.

R-DFPN-1 and R-DFPN-2 are designed to evaluate the effect of DFPN. Compared with FPN, DFPN has better feature liquidity and reusability, which is helpful to the detection. R-DFPN-2 uses DFPN and leads to 2.2% performance improvement over R-DFPN-1 (Recall improved 2.1%, Precision 2.2%).

The advantage of using rotation anchor strategy is that side effects caused by non-maximum suppression can be avoided, and it can provide higher quality proposals. The main difference between the R²CNN and R-DFPN-1 methods is the rotation anchor and pool size. Although R²CNN uses multi-scale pool size, the R²CNN Recall is still 1.8% lower than R-DFPN-1. This shows that using rotation anchors has a great effect on the Recall.

ROI Align is to solve the problem of feature misalignment. Misaligned features have a huge impact on narrow ships. In order to better preserve the semantic and spatial information of the ship, we crop the horizontal circumscribed rectangle of the proposal, meanwhile we need to regress its horizontal circumscribed rectangle at the second stage to match the feature information exactly. Compared with R-DFPN-3, R-DFPN-4 has a great improvement on Recall and Precision, and has achieved the highest F-measure of 89.6%.

We use a different pool size in order to reduce the distortion caused by interpolation. R-DFPN-2 and R-DFPN-3 show that using multiple pooled sizes has a certain promotion effect.

All detection methods are tested on the GeForce GTX 1080 (8 GB Memory). The time required for training and testing is shown in Table 2. It can be seen that the method we proposed ensures the performance improvement while keeping the training and testing time at a relatively fast level.

Table 2. Training and testing time for each method.

| Method | Faster | FPN | RRPN | R2CNN | R-DFPN-1 | R-DFPN-2 | R-DFPN-3 | R-DFPN-4 |
|--------|--------|-----|------|-------|----------|----------|----------|----------|
| Train | 0.34s | 0.5s | 0.85s | 0.5s | 0.78s | 0.78s | 1.15s | 1.15s |
| Test | 0.1s | 0.17s | 0.35s | 0.17s | 0.3s | 0.3s | 0.38s | 0.38s |

Different Recall and Precision can be obtained by changing the confidence score threshold of detection. Figure 9 plots the performance curves of different methods. The intersection of the dotted line with the other curves is the equilibrium point for each method. The location of the equilibrium point can compare the performance of the method. R-DFPN-4 (green curve) clearly has the best equilibrium point.

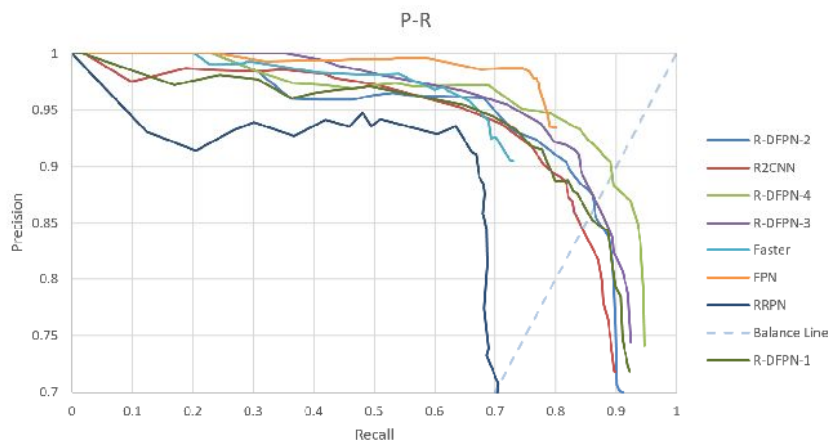

**Figure 9.** The P-R curves of different methods.

## 4. Discussion

By comparing and analyzing many groups of experiments, the validity of the proposed structure is verified. R-DFPN offers superior performance in both multi-scale and high-density objects. However, we can see from Table 1 that the Precision of R-DFPN is not the highest, behind the traditional method. Through the observation of the test results, we attribute the reasons to two points: false alarm and misjudgment.

### 4.1. False Alarm

The aspect ratio is a major feature of the ship, and the testing framework learns this. But ships tend to dock in a complex scene, such as the port, or naval base. These scenes often contain objects with similar aspect ratios, such as roofs, container pile, and docked docks. These disturbances will cause false alarms on the detector as shown in Figure 10.

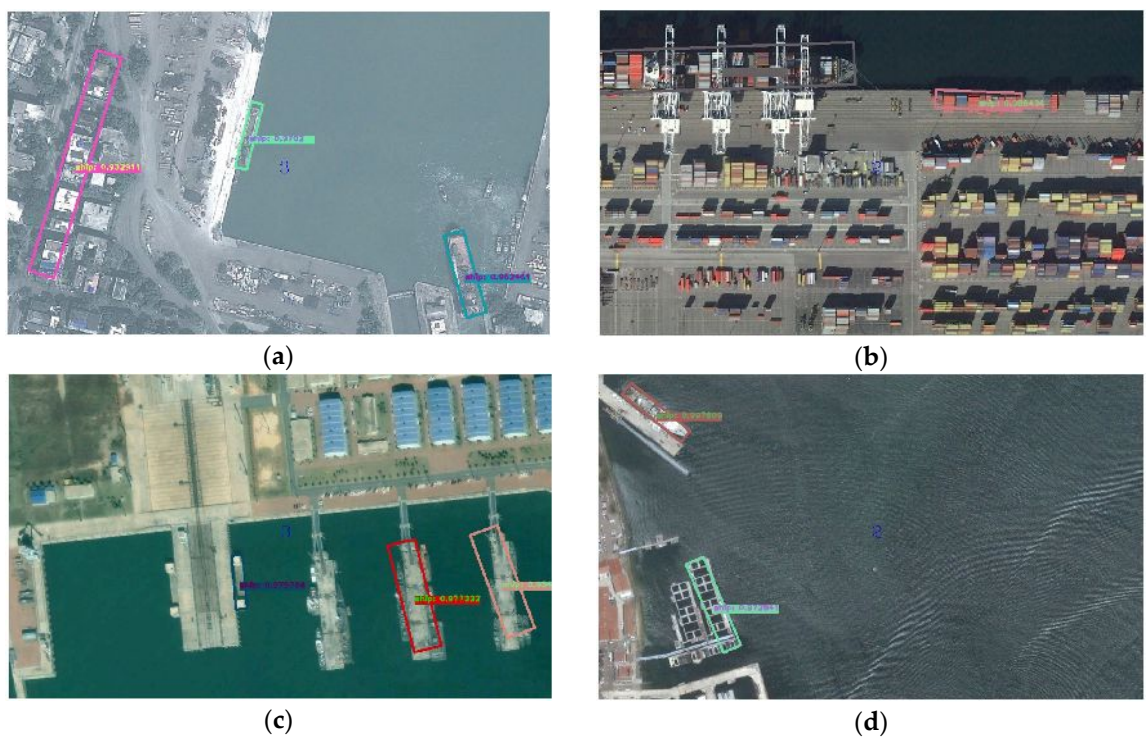

**Figure 10.** False alarms caused by different disturbances. (**a**) Roofs; (**b**) Container pile; (**c**) Docked docks; (**d**) Floating objects.

Although DFPN fully reuses feature information, it is still not enough to eliminate the false alarm effect. Sea-land segmentation can be a better solution to this problem, but limited by its segmentation accuracy. The introduction of generative adversarial nets may be another idea by training a discriminator to enhance the detector's ability to recognize.

### 4.2. Misjudgment

Misjudgment is another important reason for the low Precision. The sensitive relationship between IoU overlap and ship angle often leads to miscarriage of justice. For example, for a ship with aspect ratio of 1: 7, the IoU is only 0.38 when the angles differ by 15 degrees. When the confidence score threshold is 0.55, both targets in Figure 11 will be misjudged as not detected. This is not fair compared to traditional methods. In order to make the comparison more reasonable, we can calculate the IoU using the circumscribed rectangle of the rotating rectangle, as shown in Figure 11 (b).

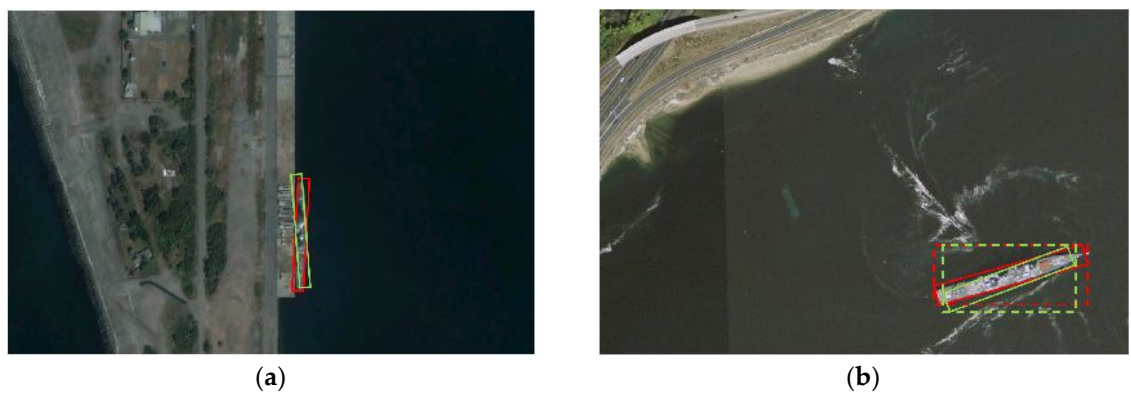

**Figure 11.** The sensitive relationship between IoU overlap and ship angle. The red box is ground-truth, and the green box is the test result.

Through this new evaluation criteria, we can recalculate the indicators, as shown in Table 3. We can see that Recall and Precision in the rotation region method all have obvious improvement.

**Table 3.** Results under the new evaluation criteria.

| Detection Method | R (%) | P (%) | F (%) |
| --- | --- | --- | --- |
| Faster | 62.7 | 96.6 | 76.0 |
| FPN | 75.5 | **97.7** | 85.2 |
| RRPN | 73.4 | 75.1 | 74.2 |
| R²CNN | 84.2 | 90.8 | 87.4 |
| R-DFPN | **90.5** | 94.1 | **92.3** |

## 5. Conclusions

In this paper, we propose a multi-scale rotation region detection method which can handle different complex scenes, detect intensive objects and reduce redundant detection region. Many novel structures are designed in this model. For example, DFPN has been designed to enhance the propagation and reuse of features. Then we use rotation anchors to improve the quality of the proposals. In addition, multi-scale ROI Align approach is adopted to fully preserve the semantic and spatial information of features. It should be noted that the regression box at second stage and ROI are both horizontal circumscribed rectangle of the proposals. Experiments show that R-DFPN has the state-of-the-art performance in the ship detection of complex scenes, especially in the task of detecting densely arranged ships.

Despite the best performance, there are still some problems. More false alarms resulted in a much lower Precision for R-DFPN than Faster-RCNN and FPN. We need to explore how to effectively reduce false alarms in the future.

**Acknowledgments:** The work is supported by the National Natural Science Foundation of China under Grants 41501485. The authors would like to thank all the colleagues in the lab, who generously provided their image dataset with the ground truth. The authors would also like to thank the anonymous reviewers for their very competent comments and helpful suggestions.

**Author Contributions:** Xue Yang and Jirui Yang conceived and designed the experiments; Xue Yang performed the experiments; Xue Yang and Jirui Yang analyzed the data; Kun Fu, Xian Sun and Menglong Yan contributed materials; Xue Yang wrote the paper. Hao Sun and Zhi Guo supervised the study and reviewed this paper.